\documentclass[11pt,a4paper]{article}
\usepackage[hyperref]{acl2018}
\usepackage{times}
\usepackage{latexsym}
\usepackage{amsmath}
\usepackage{graphicx}
\usepackage{url}
\usepackage{amssymb}
\usepackage{bm}
\usepackage{enumitem}
\usepackage{scrextend}
\usepackage{hhline}
\usepackage[ruled]{algorithm2e}

\aclfinalcopy % Uncomment this line for the final submission
\def\aclpaperid{21} %  Enter the acl Paper ID here

\DeclareSymbolFont{boldletters}{OT1}{cmr}{bx}{n}
\DeclareSymbolFontAlphabet{\mathcal}{symbols}
\DeclareMathSymbol{x}{\mathalpha}{boldletters}{`x}
\DeclareMathSymbol{c}{\mathalpha}{boldletters}{`c}
\DeclareMathSymbol{z}{\mathalpha}{boldletters}{`z}
\DeclareMathSymbol{a}{\mathalpha}{boldletters}{`a}
\DeclareMathSymbol{R}{\mathalpha}{symbols}{`R}
\DeclareMathSymbol{F}{\mathalpha}{symbols}{`F}
\DeclareMathSymbol{G}{\mathalpha}{symbols}{`G}

\renewcommand\vec[1]{\overrightarrow{#1}}
\newcommand\cev[1]{\overleftarrow{#1}}

\newcommand{\recog}{{\mathchoice{}{}{\scriptscriptstyle}{}R}}

\newcommand{\decode}{{\mathchoice{}{}{\scriptscriptstyle}{} F^d}}

\title{Zero-Shot Dialog Generation with Cross-Domain Latent Actions}

\author{Tiancheng Zhao and Maxine Eskenazi \\
 Language Technologies Institute \\
  Carnegie Mellon University \\
  Pittsburgh, Pennsylvania, USA \\
  {\tt \{tianchez, max+\}@cs.cmu.edu} \\}

\date{}

\begin{document}
\maketitle
\begin{abstract}
    This paper introduces \textit{zero-shot dialog generation} (ZSDG), as a step towards neural dialog systems that can instantly generalize to new situations with minimal data. ZSDG enables an end-to-end generative dialog system to generalize to a new domain for which only a domain description is provided and no training dialogs are available. Then a novel learning framework, Action Matching, is proposed. This algorithm can learn a cross-domain embedding space that models the semantics of dialog responses which, in turn, lets a neural dialog generation model generalize to new domains. We evaluate our methods on a new synthetic dialog dataset, and an existing human-human dialog dataset. Results show that our method has superior performance in learning dialog models that rapidly adapt their behavior to new domains and suggests promising future research.\footnote{Code and data are avaliable at \url{https://github.com/snakeztc/NeuralDialog-ZSDG}}  
\end{abstract}

\section{Introduction}
The generative end-to-end dialog model (GEDM) is one of the most powerful methods of learning dialog agents from raw conversational data in both chat-oriented and task-oriented domains~\cite{serban2016hierarchical,wen2016network,zhao2017generative}. Its base model is an encoder-decoder network~\cite{cho2014learning} that uses an encoder network to encode the dialog context and generate the next response via a decoder network. Yet prior work in GEDMs has overlooked an important issue, i.e. the data scarcity problem. In fact, the data scarcity problem is extremely common in most dialog applications due to the wide range of potential domains that dialog systems can be applied to. To the best of our knowledge, current GEDMs are data-hungry and have only been successfully applied to domains with abundant training material. This limitation prohibits the possibility of using the GEDMs for rapid prototyping in new domains and is only useful for domains with large datasets. 

The key idea of this paper lies in developing domain descriptions that can capture domain-specific information and a new type of GEDM model that can generalize to a new domain based on the domain description. Humans exhibit incredible efficiency in achieving this type of adaptation. Imagine that a customer service agent in the shoe department is transferred to the clothing department. After reading some relevant instructions and documentation, this agent can immediately begin to deal with clothes-related calls without the need for any example dialogs. We also argue that it is more efficient and natural for domain experts to express their knowledge in terms of domain descriptions rather than example dialogs. This is because creating example dialogs involves writing down imagined dialog exchanges that can be shared across multiple domains and are not relevant to the unique proprieties of a specific domain. However, current state-of-the-art GEDMs are not designed to incorporate such knowledge and are therefore incapable of adapting its behavior to unseen domains.

This paper introduces the use of \textit{zero-shot dialog generation} (ZSDG) in order to enable GEDMs to generalize to unseen situations using minimal dialog data. Building on zero-shot classification~\cite{palatucci2009zero}, we formalize ZSDG as a learning problem where the training data contains dialog data from source domains along with domain descriptions from both the source and target domains. Then at testing time, ZSDG models are evaluated on the target domain, where no training dialogs were available. We approach ZSDG by first discovering a dialog policy network that can be shared between the source and target domains. The output from this policy is distributed vectors which are referred to as \textit{latent actions}. Then, in order to transform the latent actions from any domain back to natural language utterances, a novel Action Matching (AM) algorithm is proposed that learns a cross-domain latent action space that models the semantics of dialog responses. This in turns enables the GEDM to generate responses in the target domains even when it has never observed full dialogs in them. 

Finally the proposed methods and baselines are evaluated on two dialog datasets. The first one is a new synthetic dialog dataset generated by SimDial, which was developed for this study. SimDial enables us to easily generate task-oriented dialogs in a large number of domains, and provides a test bed to evaluate different ZSDG approaches. We further test our methods on a recently released multi-domain human-human corpus~\cite{eric2017key} to validate whether performance can generalize to real-world conversations. Experimental results show that our methods are effective in incorporating knowledge from domain descriptions and achieve strong ZSDG performance. 

%We conclude by highlighting solving this objective will not only offer us more capable dialog systems, but also will advance our understanding in building AI systems that resemble closer to human's incredible learning ability. 

\section{Related Work}
Perhaps the most closely related topic is zero-shot learning (ZSL) for classification~\cite{larochelle2008zero}, which has focused on classifying unseen labels. A common approach is to represent the labels as attributes instead of class indexes~\cite{palatucci2009zero}. As a result, at test time, the model can first predict the semantic attributes in the input, then make the final prediction by comparing the predicted attributes with the candidate labels' attributes. More recent work~\cite{socher2013zero,romera2015embarrassingly} improved on this idea by learning parametric models, e.g. neural networks, to map the label and input data into a joint embedding space and then make predictions. Besides classification, prior art has explored the notion of task generalization in robotics, so that a robot can execute a new task that was not mentioned in training~\cite{oh2017zero,duan2017one}. In this case, a task is described by a demonstration or a sequence of instructions, and the system needs to learn to break down the instructions into previously learned skills. Also generating out-of-vocabulary (OOV) words from recurrent neural networks (RNNs) can be seen as a form of ZSL, where the OOV words are unseen labels. Prior work has used delexicalized tags~\cite{zhao2017generative} and copy-mechanism~\cite{gu2016incorporating,merity2016pointer,elsahar2018zero} to enable RNN output words that are not in its vocabulary.

Finally, ZSL has been applied to individual components in the dialog system pipeline. Chen et al.~\cite{chen2016zero} developed an intent classifier that can predict new intent labels that are not included in the training data. Bapna et al.~\cite{bapna2017towards} extended that idea to the slot-filling module to track novel slot types. Both papers leverage a natural language description for the label (intent or slot-type) in order to learn a semantic embedding of the label space. Then, given any new labels, the model can still make predictions. There has also been extensive work on learning domain-adaptable dialog policy by first training a dialog policy on previous domains and testing the policy on a new domain. Gasic et al.~\cite{gasic2014gaussian} used the Gaussian Process with cross-domain kernel functions. The resulting policy can leverage experience from other domains to make educated decisions in a new one. 

In summary, past ZSL research in the dialog domain has mostly focused on the individual modules in a pipeline-based dialog system. We believe our proposal is the first step in exploring the notion of adapting an entire end-to-end dialog system to new domains for domain generalization.

\section{Problem Formulation}
\label{sec:overview}
We begin by formalizing zero-shot dialog generation (ZSDG). Generative dialog models take a dialog context $c$ as input and then generate the next response $x$. ZSDG uses the term \textit{domain} to describe the difference between training and testing data. Let $D=D_s \bigcup D_t$ be a set of domains, where $D_s$ is a set of source domains, $D_t$ is a set of target domains and $D_s \cap D_t = \emptyset$. During training, we are given a set of samples $\{c^{(n)}, x^{(n)}, d^{(n)}\} \sim p_\text{source}(c, x, d)$ drawn from the \textit{source domains}. During testing, a ZSDG model will be given a dialog context $c$ and a domain $d$ drawn from the \textit{target domains} and must generate the correct response $x$. Moreover, ZSDG assumes that every domain $d$ has its own domain description $\phi(d)$ that is available at training for both source and target domains. The primary goal is to learn a generative dialog model $\mathcal{F}: C\times D \rightarrow X$ that can perform well in a target domain, by relating the unseen target domain description to the seen descriptions of the source domains. Our secondary goal is that $\mathcal{F}$ should perform similarly to a model that is designed to operate solely in the source domains. In short, the problem of ZSDG can be summarized as:
\begin{align*}
    \text{Train Data: } & \{c, x, d\} \sim p_\text{source}(c, x, d)\\
                        & \{\phi(d)\}, d \in D \\
    \text{Test Data: }  & \{c, x, d\} \sim p_\text{target}(c, x, d) \\
    \text{Goal: }       &\mathcal{F}: C\times D \rightarrow X
\end{align*}

%One natural question is why dialog generation cannot be included in the standard ZSL classification framework. The foremost issue is that unlike classification which has a finite set of unseen labels, there are an infinite number of potential responses, so that it becomes intractable to iterate through all potential responses and predict the nearest match. In fact, a generative dialog model needs to generate the response word-by-word, which is inherently different from classification. Second of all, the notion of ``unseen classes'' is invalid for dialog response generation, since responses are not categorical variables. Therefore a new notion of training and testing discrepancy needs to be developed, which is addressed by ``domains'' in our ZSDG formalism.

\section{Proposed Method}
\subsection{Seed Responses as Domain Descriptions}
The design of the domain description $\phi$ is a crucial factor that decides whether robust performance in the target domains is achievable. This paper proposes \textit{seed response} (SR) as a general-purpose domain description that can readily be applied to different dialog domains. SR needs for the developers to provide a list of example responses that the model can generate in this domain. SR's assumption is that a dialog model can discover analogies between responses from different domains, so that its dialog policy trained on source domains can be reused in the target domain. Without losing generality, $\text{SR}_d$ defines $\phi(d)$ as $\{x^{(i)}, a^{(i)}, d\}_\text{seed}$ for domain $d$, where $x$ is a seed response and $a$ is its annotations. Annotations are salient features that help the system in infer the relationship amongst responses from different domains. This may be difficult to achieve using only words in $x$, e.g. two domains with distinct word distributions. For example, in a task-oriented weather domain, a seed response can be: \textit{The weather in New York is raining} and the annotation is a semantic frame that contains domain general dialog acts and slot arguments, i.e. \textit{[Inform, loc=New York, type=rain]}. The number of seed responses is often much smaller than the number of potential responses in the domain so it is best for SR to cover more responses that are unique to this domain. SRs assume that there is a discourse-level pattern that can be shared between the source and target domains, so that a system only needs sentence-level knowledge to adapt to the target. This assumption holds in many slot-filling dialog domains and it is easy to provide utterances in the target domain that are analogies to the ones from the source domains.

\subsection{Action Matching Encoder-Decoder}
\begin{figure}[ht]
    \centering
    \includegraphics[width=0.44\textwidth]{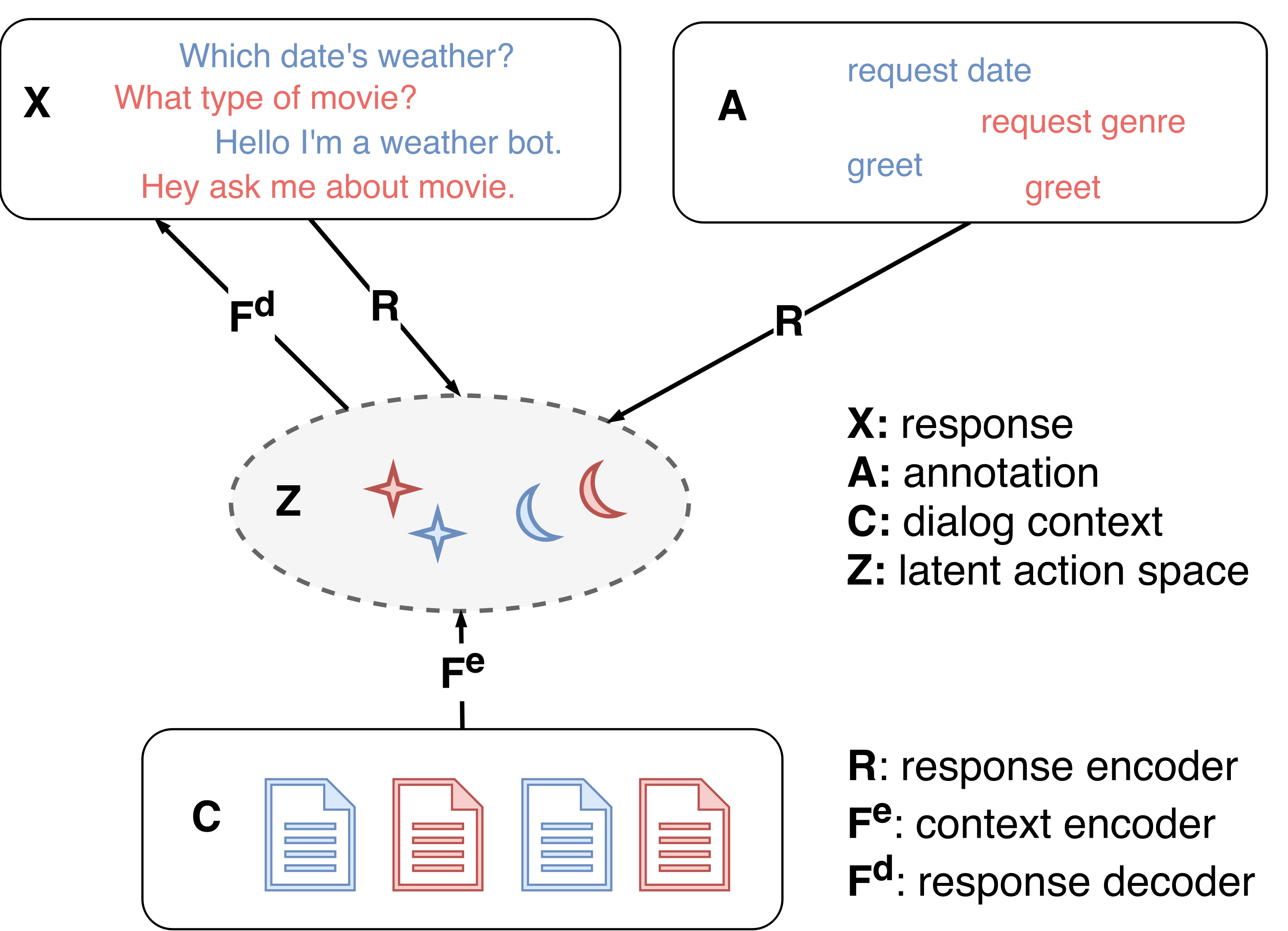}
    \caption{An overview of our Action Matching framework that looks for a latent action space $Z$ shared by the response, annotation and predicted latent action from $F^e$.}
    \label{fig:space}
\end{figure}
Figure~\ref{fig:space} shows an overview of the model we use to tackle ZSDG. The base model is a standard encoder-decoder $F$ where an encoder $F^e$ maps $c$ and $d$ into a distributed representation $z_c=F^e(c,d)$ and the decoder $F^d$ generates the response $x$ given $z_c$. We denote the embedding space that $z_c$ resides in as the \textit{latent action} space. We follow the KB-as-an-environment approach~\cite{zhao2016towards} where the generated $x$ include both system verbal utterances and API queries that interface with back-end databases. This base model has been proven to be effective in human interactive evaluation for task-oriented dialogs~\cite{zhao2017generative}.

We have two high-level goals: (1) learn a cross-domain $F$ that can be reused in all source domains and potentially shared with target domains as well. (2) create a mechanism to incorporate knowledge from the domain descriptions into $F$ so that it can generate novel responses when tested on the target domains. To achieve the first goal, we combine $c$ and $d$ by appending $d$ as a special word token at the beginning of every utterance in $c$. This simple approach performs well and enables the context encoder to take the domain into account when processing later word tokens. Also, this context domain integration can easily scale to dealing with a large number of domains. Then we encourage $F$ to discover reusable dialog policy by training the same encoder decoder on dialog data generated from multiple source domains at the same time, which is a form
 of multi-task learning~\cite{collobert2008unified}. We achieve the second goal by projecting the response $x$ from all domains into the same latent action space $Z$. Since $x$ alone may not be sufficient to infer its semantics, we rely on their annotations $a$ to learn meaningful semantic representations. Let $z_x$ and $z_a$ be the projected latent actions from $x$ and $a$. Our method encourages $z^{d_1}_{x_1} \approx z^{d_2}_{x_2}$ when $z^{d_1}_{a_1} \approx z^{d_2}_{a_2}$. Moreover, for a given $z$ from any domain, we ensure that the decoder $F^d$ can generate the corresponding response $x$ by training on both $\text{SR}_d$ for $d \in D$ and source dialogs. 

Specifically, we propose the Action Matching (AM) training procedure. We first introduce a recognition network $R$ that can encode $x$ and $a$ into $z_x=R(x, d)$ and $z_a=R(a, d)$ respectively. During training, the model receives two types of data. The first type is domain description data in the form of $\{x, a, d\}_{seed}$ for each domain. The second type of data is source domain dialog data in the form of $\{c, x, d\}$.  For the first type of data, we update the parameters in $\recog$ and $F^d$ by minimizing the following loss function:
\begin{align}
\begin{split}
    \mathcal{L}_\text{dd}(F^d, R) =&  -\log p_\decode(x|\recog(a, d)) \\ 
                & + \lambda \mathbb{D}[\recog(x, d) \| \recog(a, d)]
\end{split}
\end{align}
where $\lambda$ is a constant hyperparameter and $\mathbb{D}$ is a distance function, e.g. mean square error (MSE), that measures the closeness of two input vectors. The first term in $\mathcal{L}_\text{dd}$ trains the decoder $F^d$ to generate the response $x$ given $z_a=R(a, d)$ from all domains. The second term in $\mathcal{L}_\text{dd}$ enforces the recognition network $R$ to encode a response and its annotation to nearby vectors in the latent action space from all domains, i.e. $z^d_x \approx z^d_a$ for $d \in D$. 

Moreover, just optimizing $\mathcal{L}_\text{dd}$ does not ensure that the $z_c$ predicted by the encoder $F^e$ will be related to the $z_x$ or $z_a$ encoded by the recognition network $\recog$. So when we receive the second type of data (source dialogs), we add a second term to the standard maximum likelihood objective to train $F$ and $\recog$.  
\begin{align}
\begin{split}
    \mathcal{L}_\text{dialog}(F, R) = & -\log p_\decode(x|F^e(c, d)) \\
                                & + \lambda \mathbb{D}(\recog(x, d) \| F^e(c, d))
\end{split}
\end{align}
The second term in $\mathcal{L}_\text{dialog}$ completes the loop by encouraging $z^d_c \approx z^d_x$, which resembles the regularization term used in variational autoencoders~\cite{kingma2013auto}. Assuming that annotation $a$ provides a domain-agnostic semantic representation of $x$, then $F$ trained on source domains can begin to operate in the target domains as well. During training, our AM algorithm alternates between these two types of data and optimizes $\mathcal{L}_\text{dd}$ or $\mathcal{L}_\text{dialog}$ accordingly. The resulting models effectively learn a latent action space that is shared by the the response annotation $a$, response $x$ and predicted latent action based on $c$ in all domains. AM training is summarized in Algorithm~\ref{al:am}.
\begin{algorithm}[ht]
\label{al:am}
\caption{Action Matching Training}
\SetAlgoLined
Initialize weights of $F^e$, $F^d$, $R$;\\
Data = $\{c, x, d\} \bigcup \{x, a, d\}_{seed}$ \\
 \While{batch $\sim$ Data}{
  \eIf{$\text{batch in the form } \{c, x, d\}$}{
   Backpropagate loss $\mathcal{L}_\text{dialog}$
   }{
   Backpropagate loss $\mathcal{L}_\text{dd}$
  }
 }
\end{algorithm}

\subsection{Architecture Details}
\begin{figure*}[ht]
    \centering
    \includegraphics[width=16cm]{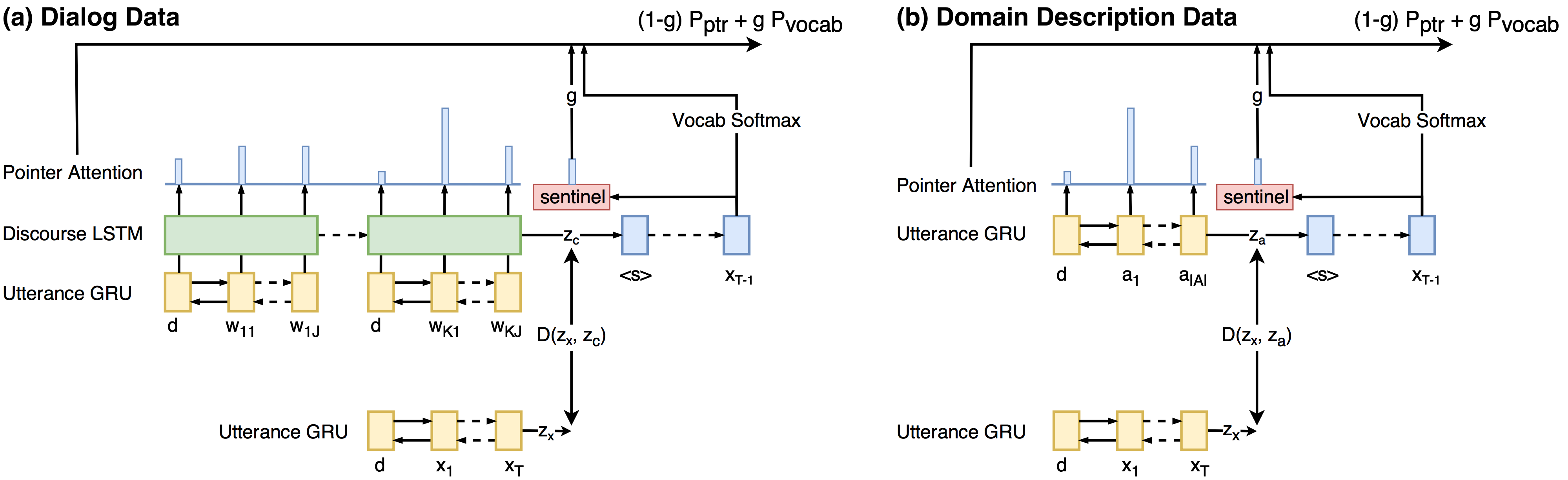}
    \caption{Visual illustration of our AM encoder decoder with copy mechanism~\cite{merity2016pointer}. Note that AM can also be used with RNN decoders without the copy functionality.}
    \label{fig:model}
\end{figure*}
We implement an AMED for later experiments as follows:

\textbf{Distance Functions}: In this study, we assume that the latent actions are deterministic distributed vectors. Thus MSE is used: $\mathbb{D}(z, \hat{z})=\frac{1}{L}\sum_l^L(z_l-\hat{z}_l)^2$, where $L$ is the dimension size of the latent actions. Also, $L_\text{dialog}$ and $L_\text{dd}$ use the same distance function.

\textbf{Recognition Networks}: we use a bidirectional GRU-RNN~\cite{cho2014learning} as $\recog$ to obtain utterance-level embedding. Since both $x$ and $a$ are sequences of word tokens, we combine them with the domain tag by appending the domain tag in the beginning of the original word sequence, i.e. $\{x, d\} \text{ or } \{a, d\} = [d, w_1, ... w_J]$, where $J$ is the length of the word sequence. Then the $R$ will encode $[d, w_1, ... w_J]$ into hidden outputs in forward and backward directions, $[(\vec{h_0}, \cev{h_J}), ...(\vec{h_J}, \cev{h_0})]$. We use the concatenation of the last hidden states from each direction, i.e. $z_x \text{ or } z_a =[\vec{h_J}, \cev{h_J}]$ as utterance-level embedding for $x$ or $a$ respectively.

\textbf{Dialog Encoders}: a hierarchical recurrent encoder (HRE) is used to encode the dialog context, which handles long contexts better than non-hierarchical ones~\cite{li2015hierarchical}. HRE first uses an utterance encoder to encode every utterance in the dialog and then uses a discourse-level LSTM-RNN to encode the dialog context by taking output from the utterance encoder as input. Instead of introducing a new utterance encoder, we reuse the recognition network $R$ described above as the utterance encoder, which serves the purpose perfectly. Another advantage is that using $z_x$ predicted by $R$ as input enables the discourse-level encoder to use knowledge from latent actions as well. Our discourse-level encoder is a 1-layer LSTM-RNN~\cite{hochreiter1997long}, which takes in a list of output $[z_1, z_2 .. z_K]$ from $R$ and encodes them into $[v_1, v_2, ... v_K]$, where $K$ is the number of utterances in the context. The last hidden state $v_K$ is used as the predicted latent action $z_c$. 

\textbf{Response Decoders}: we experiment with two types of LSTM-RNN decoders. The first is an RNN decoder with an attention mechanism~\cite{luong2015effective}, enabling the decoder to dynamically look up information from the context. Specifically, we flatten the dialog context into a sequence of words $[w_{11}, ... w_{1J} ... w_{KJ}]$. Using output from the $R$ and the discourse-level LSTM-RNN, each word here is represented by $m_{kj}= h_{kj} + W_v v_k$. Let the hidden state of the decoder at step $t$ be $s_t$, then our attention mechanism computes the Softmax output via:
\begin{align}
    \alpha_{kj,t} &= \text{softmax}(m^T_{kj} \tanh(W_\alpha s_t))\\
    \widetilde{s_t} &= \sum_{kj} \alpha_{kj,t} m_{kj}\\
    p_\text{vocab}(w_t|s_t) &= \text{softmax}(\text{MLP}(s_t, \widetilde{s_t}))
\end{align}
The second type is the LSTM-RNN with a copy mechanism that can directly copy words from the context as output~\cite{gu2016incorporating}. Such a mechanism has already exhibited strong performance in task-oriented dialogs~\cite{eric2017copy} and is well suited for generating OOV word tokens~\cite{elsahar2018zero}. We implemented the Pointer Sentinel Mixture Model (PSM)~\cite{merity2016pointer} as our copy decoder. PSM defines the generation of the next word as a mixture of probabilities from either the Softmax output from the decoder LSTM or the attention Softmax for words in the context: $p(w_t|s_t) = g p_\text{vocab}(w_t|s_t) + (1-g) p_\text{ptr}(w_t|s_t)$, where $g$ is the mixture weight computed from a sentinel vector $u$ with $s_t$.
\begin{align}
    p_\text{ptr}(w_t|s_t) &= \sum_{kj \in I(w, x)}  \alpha_{kj, t} \\
    g & = \text{softmax}(u^T \tanh(W_\alpha s_i)) 
\end{align}

\section{Datasets for ZSDG}
Two dialog datasets were used for evaluation.
\subsection{SimDial Data}
We developed SimDial\footnote{\url{https://github.com/snakeztc/SimDial}}, which is a multi-domain dialog generator that can generate realistic conversations for slot-filling domains with configurable complexity. See Appendix~\ref{sec:simdial_details} for details. Compared to other synthetic dialog corpora used to test GEDMs, e.g. bAbI~\cite{dodge2015evaluating}, SimDial data is significantly more challenging. First since SimDial simulates communication noise, the dialogs that are generated can be very long (more than 50 turns) and the simulated agent can carry out error recovery strategies to correctly infer the users' goals. This challenges end-to-end models to model long dialog contexts. SimDial also simulates spoken language phenomena, e.g. self-repair, hesitation. Prior work~\cite{eshghi2017bootstrapping} has shown that this type of utterance-level noise deteriorates end-to-end dialog system performance. 

\subsubsection*{Data Details}
SimDial was used to generate dialogs for 6 domains: restaurant, movie, bus, restaurant-slot, restaurant-style and weather. For each domain, 900/100/500 dialogs were generated for training, validation and testing. On average, each dialog had 26 utterances and each utterance had 12.8 word tokens. The total vocabulary size was 651. We split the data such that the training data included dialogs from the restaurant, bus and weather domains and the test data included the restaurant, movie, restaurant-slot and restaurant style domains. This setup evaluates a ZSDG system from the following perspectives:

\textbf{Restaurant (in domain)}: evaluation on the restaurant test data checks if a dialog model is able to maintain its performance on the source domains. \textbf{Restaurant-slot (unseen slots)}: restaurant-slot has the same slot types and natural language generation (NLG) templates as the restaurant domain, but has a completely different slot vocabulary, i.e. different location names and cuisine types. Thus this is designed to evaluate a model that can generalize to unseen slot values. \textbf{Restaurant-style (unseen NLG)}: restaurant-style has the same slot type and vocabulary as restaurant, but its NLG templates are completely different, e.g. ``which cuisine type?'' $\rightarrow$ ``please tell me what kind of food you prefer''. This part tests whether a model can learn to adapt to generate novel utterances with similar semantics. \textbf{Movie (new domain)}: movie has completely different NLG templates and structure and shares few common traits with the source domains at the surface level. Movie is the hardest task in the SimDial data, which challenges a model to correctly generate next responses that are semantically different from the ones in source domains.

Finally, we obtain SRs as domain descriptions by randomly selecting 100 unique utterances from each domain. The response annotation is a response's internal semantic frame used by the SimDial generator. For example, ``I believe you said Boston. Where are you going?'' $\rightarrow$ [implicit-confirm loc=Boston; request location]. 

\subsection{Stanford Multi-Domain Dialog Data}
The second dataset is the Stanford multi-domain dialog (SMD) dataset~\cite{eric2017key} of 3031 human-human dialogs in three domains: weather, navigation and scheduling. One speaker plays the role of a driver. The other plays the car's AI assistant and talks to the driver to complete tasks, e.g. setting directions on a GPS. Average dialog length is 5.25 utterances; vocabulary size is 1601. We use SMD to validate whether our proposed methods generalize to human-generated dialogs. We generate SR by randomly selecting 150 unique utterances for each domain. An expert annotates the seed utterances with dialog acts and entities. For example ``All right, I've set your next dentist appointment for 10am. Anything else?'' $\rightarrow$ [ack; inform goal event=dentist appointment time=10am ; request needs]. Finally, in order to formulate a ZSDG problem, we use a leave-one-out approach with two domains as source domains and the third one as the target domain, which results in 3 possible configurations.

\section{Experiments and Results}
\label{sec:exp_setup}

The baseline models include 1. hierarchical recurrent encoder with attention decoder (+Attn)~\cite{serban2016hierarchical}. 2. hierarchical recurrent encoder with copy decoder~\cite{merity2016pointer} (+Copy), which has achieved very good performance on task-oriented dialogs~\cite{eric2017copy}. We then augment both baseline models with the proposed cross-domain AM training procedure and denote them as +Attn+AM and +Copy+AM. 

Evaluating generative dialog systems is challenging since the model can generate free-form responses. Fortunately, we have access to the internal semantic frames of the SimDial data, so we use the automatic measures used in~\cite{zhao2017generative} that employ four metrics to quantify the performance of a task-oriented dialog model. \textbf{BLEU} is the corpus-level BLEU-4 between the generated response and the reference ones~\cite{papineni2002bleu}. \textbf{Entity F$_\text{1}$} checks if a generated response contains the correct entities (slots) in the reference response. \textbf{Act F$_\text{1}$} measures whether the generated responses reflect the dialog acts in the reference responses, which compensates for BLEU's limitation of looking for exact word choices. A one-vs-rest support vector machine~\cite{scholkopf2001learning} with bi-gram features is trained to tag the dialogs in a response. \textbf{KB F$_\text{1}$} checks all the key words in a KB query that the system issues to the KB backend. Finally, we introduce $\textbf{BEAK}=\sqrt[4]{\text{bleu} \times \text{ent} \times \text{act} \times \text{kb}}$, the geometric mean of these four scores, to quantify a system's overall performance. Meanwhile, since the oracle dialog acts and KB queries are not provided in the SMD data~\cite{eric2017key}, we only report BLEU and entity F$_\text{1}$ results on SMD. 

\subsection{Main Results}
\begin{table}[ht]
\centering
\small
\begin{tabular}{p{0.07\textwidth}|p{0.07\textwidth}p{0.07\textwidth}p{0.07\textwidth}p{0.07\textwidth}} \hline
\textbf{In domain}   & +Attn & +Copy  & +Attn +AM & +Copy +AM \\ \hline
BLEU        & 59.1  & \textbf{70.4}   & 67.7  & 70.1         \\
Entity      & 69.2  & 70.5   & 74.1  & \textbf{79.9}         \\
Act         & 94.7  & 92.0   & 94.1  & \textbf{95.1}         \\
KB          & 94.7  & 96.1   & 95.2  & \textbf{97.0}         \\
BEAK        & 77.2  & 81.3   & 81.9  & \textbf{84.7}         \\ \hline
\textbf{Unseen Slot} & +Attn & +Copy  & +Attn +AM & +Copy +AM \\ \hline
BLEU        & 24.9  & 45.6     & 47.9   & \textbf{68.5}     \\
Entity      & 56.0  & 68.0     & 53.1  & \textbf{74.6}    \\
Act         & 90.9  & 91.8     & 86.0  & \textbf{94.5}    \\
KB          & 78.1  & 89.6     & 81.0  & \textbf{95.3}    \\
BEAK        & 56.1  & 71.1     & 64.8  & \textbf{82.3}     \\ \hline
\textbf{Unseen NLG} & +Attn   & +Copy  & +Attn +AM & +Copy +AM \\ \hline
BLEU        & 15.8  & 36.9    & 43.5  & \textbf{70.1}     \\
Entity      & 61.7  & 68.9    & 63.8  & \textbf{72.9}     \\
Act         & 91.5  & 92.2    & 89.3  & \textbf{95.2}    \\
KB          & 66.2  & 94.6    & 93.1  & \textbf{97.0}     \\
BEAK        & 49.3  & 65.9    & 69.3  & \textbf{82.9}    \\ \hline
\textbf{New ~domain} & +Attn  & +Copy  & +Attn +AM & +Copy +AM \\ \hline
BLEU        & 13.5  & 24.6   &  36.7  & \textbf{54.6}   \\
Entity      & 23.1  & 40.8   & 23.3   & \textbf{52.6}     \\
Act         & 82.3  & 85.5   & 84.8   & \textbf{88.5}     \\
KB          & 43.5  & 67.1   & 67.0   & \textbf{88.2}     \\
BEAK        & 32.5  & 48.8   & 46.8   & \textbf{68.8}    \\ \hline
\end{tabular}
\caption{Evaluation results on test dialogs from SimDial Data. Bold values indicate the best performance.}
\label{tbl:result_simdial}
\end{table}
Table~\ref{tbl:result_simdial} shows results on the SimDial data. Although the standard +Attn model achieves good performance in the source domains, it doesn't generalize to target domains, especially for entity F$_\text{1}$ in the unseen-slot domain, BLEU score in the unseen-NLG domain, and all new domain metrics. The +Copy model has better, although still limited, generalization to target domains. The main benefit of the +Copy model is its ability to directly copy and output words from the context, reflected in its strong entity F$_\text{1}$ in the unseen slot domain. However, +Copy can't generalize to new domains where utterances are novel, e.g. the unseen NLG or the new domain. However, our AM algorithm substantially improves performance of both decoders (Attn and Copy). Results show that the proposed AM algorithm is complementary to decoders with a copy mechanism: HRED+Copy+AM model has the best performance on all target domains. In the easier unseen-slot and unseen-NLG domains, the resulting ZSDG system achieves a BEAK of about 82, close to the in-domain BEAK performance (84.7). Even in the new domain (movie), our model achieves a BEAK of 67.2, 106\% relative improvement w.r.t +Attn and 38.8\% relative improvement w.r.t +Copy. Moreover, our AM method also improves performance on in-domain dialogs,  suggesting that AM exploits the knowledge encoded in the domain description and improves the models' generalization. 

\begin{table}[ht]
\centering
\small
\begin{tabular}{p{0.07\textwidth}|p{0.07\textwidth}p{0.07\textwidth}p{0.07\textwidth}p{0.07\textwidth}}  \hline
\textbf{Navigate} & Oracle  & +Attn  & +Copy   & +Copy +AM              \\ \hline
BLEU              & 13.4    & 0.9      & 5.4     & \textbf{5.9}                     \\
Entity            & 19.3    & 2.6      & 4.7     & \textbf{14.3}                    \\\hhline{=====}
\textbf{Weather}  & Oracle  & +Attn   & +Copy   & +Copy +AM      \\ \hline
BLEU              &  18.9    &  4.8    & 4.4     & \textbf{8.1}   \\
Entity            &  51.9   &  0.0   &  16.3   &  \textbf{31.0} \\\hhline{=====}
\textbf{Schedule} & Oracle  & +Attn  & +Copy & +Copy +AM               \\ \hline
BLEU              & 20.9     &  3.0   &  3.8  & \textbf{7.9}         \\
Entity            & 47.3     &  0.4  &  17.1  &  \textbf{36.9}       \\ \hline
\end{tabular}
\caption{Evaluation on SMD data. The bold domain title is the one that was excluded from training.}
\label{tbl:result_smd}
\end{table}
Table~\ref{tbl:result_smd} summarizes the results on the SMD data. We also report the oracle performance, obtained by training +Copy on the full dataset. The AM algorithm can  significantly improve Entity F$_\text{1}$ and BLEU from the two baseline models. +Copy+AM also achieves competitive performance in terms of Entity F$_\text{1}$ compared to the oracle scores, despite the fact that no target domain data was used in training. 

\label{sec:qual-analysis}
\begin{table*}[!ht]
\centering
\small
\begin{tabular}{p{0.13\textwidth}|p{0.18\textwidth}p{0.18\textwidth}p{0.18\textwidth}p{0.18\textwidth}} \hline
\textbf{Type}            & \textbf{Reference}                 & \textbf{+Attn}   & \textbf{+Copy}         & \textbf{+Copy+AM}                          \\ \hline
General Utts       & See you next time.                 & Goodbye.                     & See you next time.                     & See you next time.                 \\
Unseen Slots   & Do you mean romance movie?         & Do you mean Chinese food.    & Do you mean \textbf{romance} food?              & Do you mean \textbf{romance} movie?         \\
Unseen Utts         & Movie 55 is a great movie.         & Bus 12 can take you there.   & Bus \textbf{55} can take you there.    & Movie \textbf{55} is a great movie.         \\ \hline
\end{tabular}
\caption{Three types of responses and generation results (tested on the new movie domain). The text in bold is the output directly copied from the context by the copy decoder.}
\label{tbl:error}
\end{table*}

\subsection{Model Analysis}
\begin{figure}[ht]
    \centering
    \includegraphics[width=0.49\textwidth]{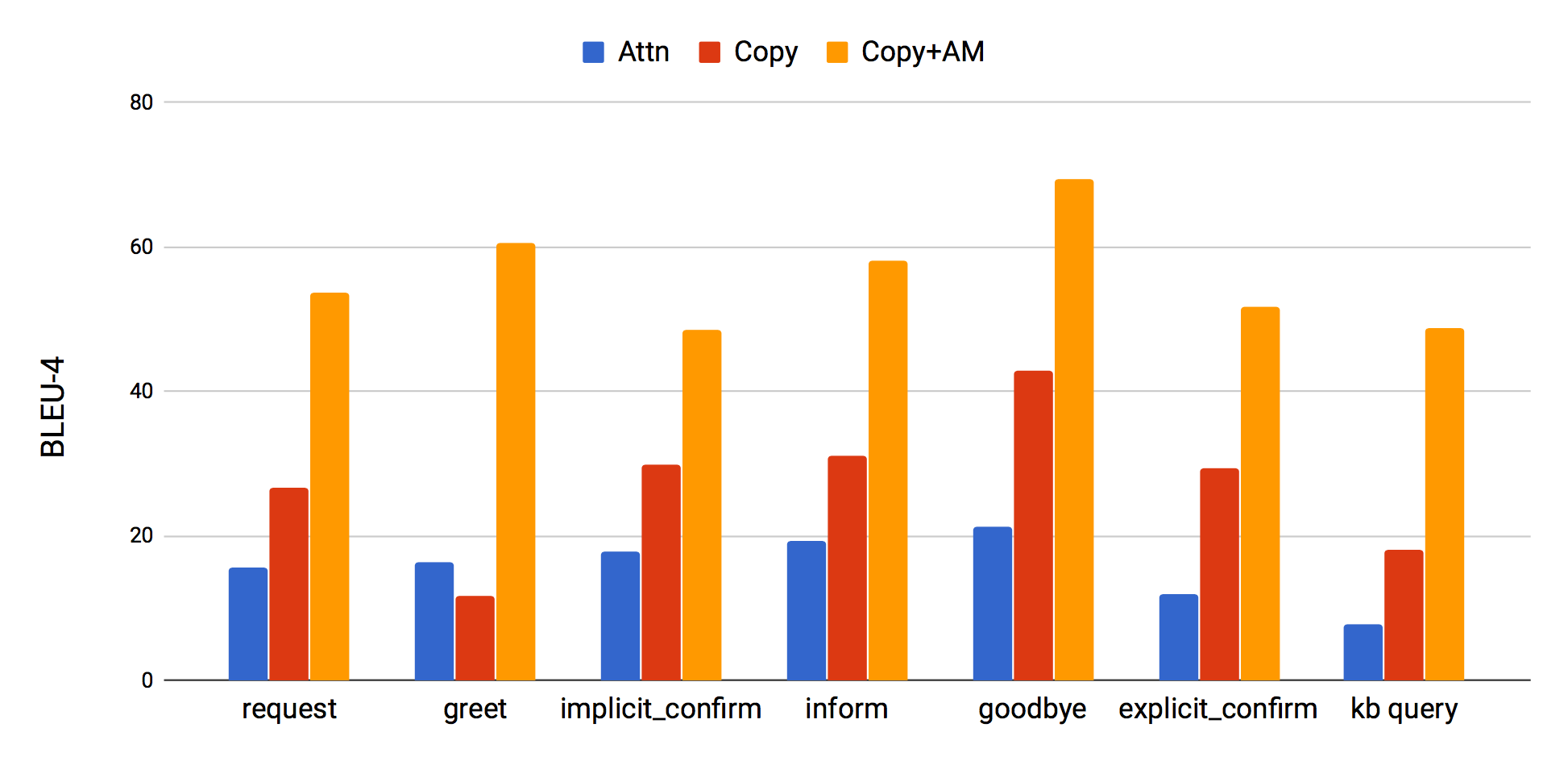}
    \caption{Breakdown BLEU scores on the new domain test set from SimDial.}
    \label{fig:breakdown}
\end{figure}

Various types of performance improvement were also studied. Figure~\ref{fig:breakdown} shows the breakdown of the BLEU score according to the dialog acts of reference responses. Models with +Copy decoder can improve performance for all dialog acts except for the \textit{greet} act, which occurs at the beginning of a dialog. In this case, the +Copy decoder has no context to copy and thus cannot generate any novel responses. This is one limitation of +Copy decoder since in real interactive testing with humans, each system utterance must be generated from the model instead of copied from the context. However, models with AM training learn to generate novel utterances based on knowledge from the SR, so +Copy+AM can generate responses at the beginning of a dialog. 

A qualitative analysis was conducted to summarize typical responses from these models. Table~\ref{tbl:error} shows three types of typical situations in the SimDial data. The first type is \textbf{general utterance} utterances, e.g. ``See you next time'' that appear in all domains. All three models correctly generate them in the ZSDG setting. The second type is utterances with \textbf{unseen slots}. For example, explicit confirm ``Do you mean xx?''. +Attn fails in this situation since the new slot values are not in its vocabulary. +Copy still performs well since it learns to copy entity-like words from the context, but the overall sentence is often incorrect, e.g. ``Do you mean romance food''. The last one is \textbf{unseen utterance} where both +Attn and +Copy fail. The two baseline models can still generate responses with correct dialog acts, but the output words are in the source domains. Only the models trained with AM are able to infer that ``Movie xx is a great movie'' serves a function similar to ``Bus xx can take you there'', and generates responses using the correct words from the target domain.

\begin{figure}[bt]
    \centering
    \includegraphics[width=0.49\textwidth]{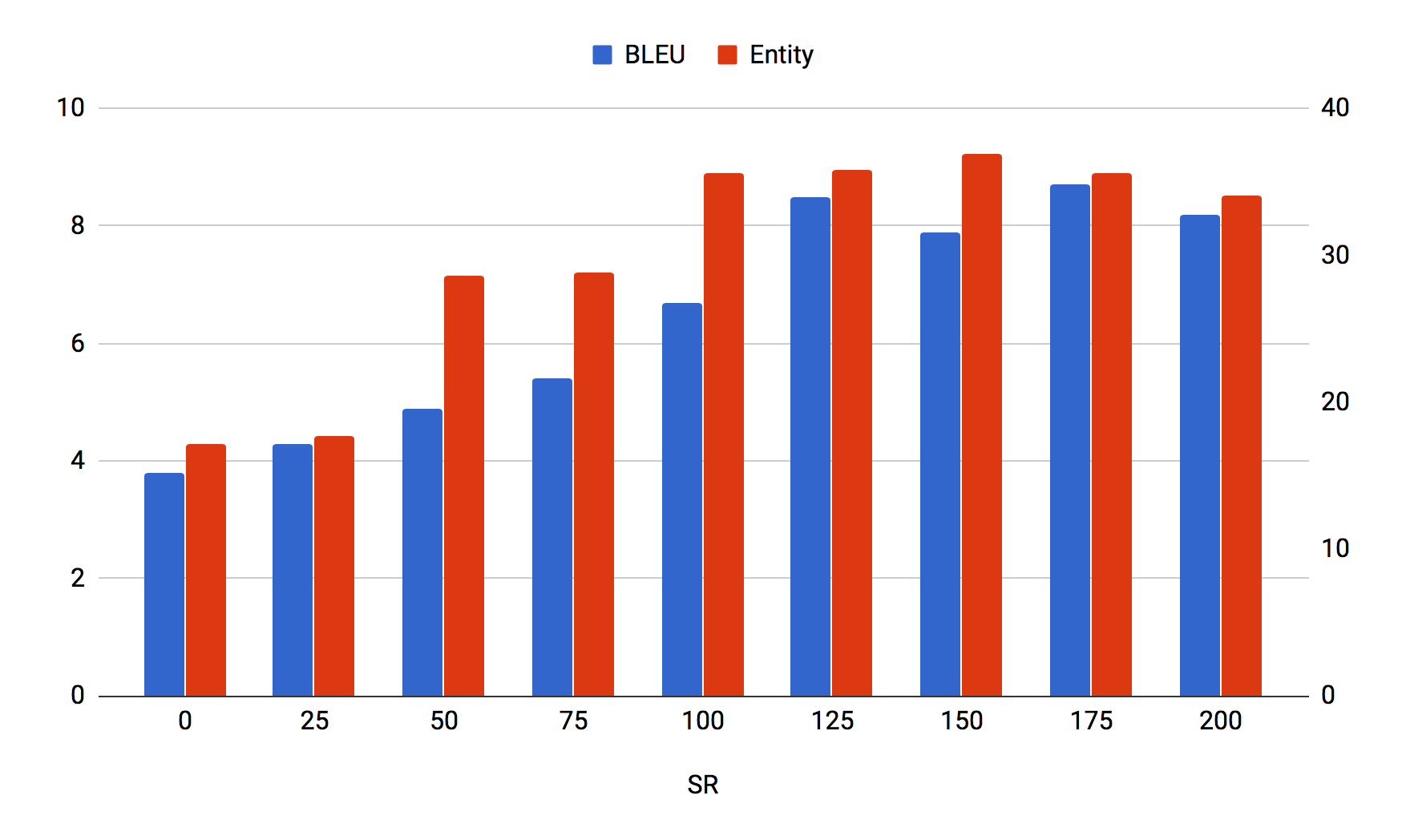}
    \caption{Performance on the schedule domain from SMD while varying the size of SR.}
    \label{fig:size}
\end{figure}

Finally we investigate how the the size of SR affects AM performance. Figure~\ref{fig:size} shows results in the SMD schedule domain. The number of seed responses varies from 0 to 200. Performance in the target domains is positively correlated with the number of seed responses. We also observe that the model achieves sufficient SR performance at 100, compared to the ones trained on all of the 200 seed responses. This suggests that the amount of seeding needed by SR is relatively small, which shows the practicality of using SR as a domain description.

\section{Conclusion and Future Work}
This paper introduces ZSDG, dealing with neural dialog systems' domain generalization ability. We formalize the ZSDG problem and propose an Action Matching framework that discovers cross-domain latent actions. We present a new simulated multi-domain dialog dataset, SimDial, to benchmark the ZSDG models. Our assessment validates the AM framework's effectiveness and the AM encoder decoders perform well in the ZSDG setting. 

ZSDG provides promising future research questions. How can we reduce the annotation cost of learning the latent alignment between actions in different domains? How can we create ZSDG for new domains where the discourse-level patterns are significantly different? What are other potential domain description formats? In summary, solving ZSDG is an important step for future general-purpose conversational agents.
% \section*{Acknowledgments}

%\bibliography{acl2018}
\bibliography{acl2018}
\bibliographystyle{acl_natbib}

\appendix
\section{Supplemental Material}
\subsection{Seed Response Creation Process}
We follow the following process to create SR in a new slot-filling domain. First, we collect seed responses (including user/system utterances, KB queries and KB responses) from each source domain and annotate them with dialog acts, entity types and entity values. Then human experts with knowledge about the target domain can write up seed responses for the target domain by drawing ideas from the sources. For example, if the source domain is restaurants and the target domain is movies. The source may contain a system utterance with its annotation: ``I believed you said Pittsburgh, what kind of food are you interested in? $\rightarrow$ \textit{[implicit-confirm, loc=Pittsburgh, request food type]}''. Then the expert can come up with a similar utterance from the target domain, e.g. ``Alright, Pittsburgh. what type of movie do you like? $\rightarrow$ \textit{[implicit-confirm, loc=Pittsburgh, request movie type]}''. In this way, our proposed AM training can leverage the annotations to match these two actions as analogies in the latent action space. Another advantage of this process is that human experts do not have to directly label whether two utterances from two domains are direct analogies; this could be ambiguous and challenging. Instead, human experts only create domain shareable annotations and leave the difficult matching problem to our models.

\subsection{Model Details}
For all experiments, we use a word embedding with size 200. The recognition network uses bidirectional GRU-RNN with hidden size 256 for each direction. The discourse-level LSTM-RNN and the decoder have hidden size 512. The models are trained with Adam~\cite{kingma2014adam} with learning rate 0.001. All weights are initialized from uniform distribution between +/-0.08. 30\% dropouts are applied at the input and output of recurrent neural networks~\cite{zaremba2014recurrent}. For decoders with copy mechanisms, we also use the pointer loss proposed in~\cite{merity2016pointer} which minimizes $-\log (g + \sum_{kj \in I(x, c)} \alpha_{kj})$, where $I$ results in all positions of output $x$ in the attention context.

\subsection{SimDial Details}
 To generate data, SimDial expects: a \textit{domain specification} (DS) and a \textit{complexity specification} (CS). DS defines the content of the domain, e.g. restaurant or weather and CS defines complexity of dialogs for three aspects: 
 
\textbf{Environmental:} the complexity level of the communication channel, e.g. automatic speech recognition (ASR) error rate. 

\textbf{Propositional:} the complexity level in the propositional content of user utterances, e.g. the chance that a user will give multiple slot values in one utterance. 

\textbf{Interaction:} the complexity level in terms of real-time interaction, e.g. the percentage of self-repair.
\label{sec:simdial_details}

The following is an example dialog generated for the restaurant domain with all of the above complexity turned on. The number at the end of the speaker turn indicates ASR confidence. 
\begin{table}[ht]
    \centering
    \small
    \begin{tabular}{p{0.45\textwidth}}
        \textbf{Domain: Restaurant} \\
        \textbf{Sys}: Welcome to the restaurant recommendation system. What do you need?\\
        \textbf{User (0.83)}: Hello robot. I am looking for a restaurant.\\
        \textbf{Sys}: Which city are you interested in?\\
        \textbf{User (0.75)}: I am in San Jose. And Korean. Uhm sorry, I am interested in Chinese food.\\
        \textbf{Sys}: I believe you said San Jose. I believe you said Chinese.\\
        \textbf{Sys}: QUERY goal: restaurant food-pref: Chinese location: San Jose\\
        \textbf{KB}: RET restaurant: 17\\
        \textbf{Sys}: Restaurant 17 is a good choice. What else can I do for you?\\
        \textbf{User (0.6)}: Not done yet. Is that restaurant open now?\\
        \textbf{Sys}: QUERY goal: open restaurant: 17 food-pref: Chinese location: San Jose\\
        \textbf{KB}: RET open: yes\\
        \textbf{Sys}: Yes, it is open right now. What else?\\
        \textbf{User (0.69)}: No more questions. Thanks. \\
        \textbf{Sys}: Bye. \\
    \end{tabular}
    \caption{An example dialog generated from SimDial.}
    \label{tab:my_label}
\end{table}

\end{document}